\documentclass{article}

\usepackage{arxiv}

\usepackage[utf8]{inputenc} 
\usepackage[T1]{fontenc}    
\usepackage{hyperref}       
\usepackage{url}            
\usepackage{booktabs}       
\usepackage{amsfonts}       
\usepackage{nicefrac}       
\usepackage{microtype}      
\usepackage{lipsum}		
\usepackage{graphicx}
\usepackage[square,numbers]{natbib}
\usepackage{doi}

\usepackage{times}
\usepackage{epsfig}
\usepackage{graphicx}
\usepackage{amsmath}
\usepackage{amssymb}
\usepackage{booktabs}
\usepackage{algorithm,algorithmic}
\usepackage{multirow}
\usepackage{subfigure}
\usepackage[symbol]{footmisc}

\title{Simultaneous Perturbation Method for Multi-Task Weight Optimization in One-Shot Meta-Learning}

\author{{Andrei Boiarov} \\
	SIT Rolos\\
	Schaffhausen, Switzerland \\
	\texttt{andrei.boiarov@sit.team} \\
	\And
	{Kostiantyn Khabarlak} \\
	Dnipro University of Technology\\
	Dnipro, Ukraine\\
	\texttt{habarlack@gmail.com} \\
	\And
	{Igor Yastrebov} \\
	OTP Bank\\
	Budapest, Hungary\\
}

\date{}

\renewcommand{\shorttitle}{Simultaneous Perturbation Optimization in One-Shot Meta-Learning}

\hypersetup{
pdftitle={Simultaneous Perturbation Optimization in One-Shot Meta-Learning},
pdfauthor={Andrei Boiarov, Konstantin Khabarlak, Igor Yastrebov},
pdfkeywords={Meta-Learning, Multi-Task Learning, One-Shot Learning, Stochastic Approximation, Deep Learning},
}

\begin{document}
\maketitle

\begin{abstract}
	Meta-learning methods aim to build learning algorithms capable of quickly adapting to new tasks in low-data regime. One of the most difficult benchmarks of such algorithms is a one-shot learning problem. In this setting many algorithms face uncertainties associated with limited amount of training samples, which may result in overfitting. This problem can be resolved by providing additional information to the model. One of the most efficient ways to do this is multi-task learning. In this paper we investigate the modification of a standard meta-learning pipeline. The proposed method simultaneously utilizes information from several meta-training tasks in a common loss function. The impact of these tasks in the loss function is controlled by a per task weight. Proper optimization of the weights can have big influence on training and the final quality of the model. We propose and investigate the use of methods from the family of Simultaneous Perturbation Stochastic Approximation (SPSA) for optimization of meta-train tasks weights. We also demonstrate superiority of stochastic approximation in comparison to gradient-based method. The proposed Multi-Task Modification can be applied to almost all meta-learning methods. We study applications of this modification on Model-Agnostic Meta-Learning and Prototypical Network algorithms on CIFAR-FS, FC100, miniImageNet and tieredImageNet one-shot learning benchmarks. During these experiments Multi-Task Modification has demonstrated improvement over original methods. SPSA-Tracking algorithm first adapted in this paper for multi-task weight optimization shows the largest accuracy boost that is competitive to the state-of-the-art meta-learning methods. Our code is available online\footnote[1]{https://github.com/andrewbo29/mtm-meta-learning-sa}.
\end{abstract}


\section{Introduction}

One of the most important hallmarks of human intelligence is the ability to learn new concepts from few examples. However, despite significant progress in deep learning in a variety of fields in recent years limitations of deep learning approaches still exist in many practical applications where labeled data is intrinsically rare or expensive. For machine learning systems it is difficult to adapt to new concepts fast with very little supervision. That raises attention to the challenging few-shot learning problem and its most difficult setting: one-shot learning. One of the promising areas of research in recent years was to tackle this problem using the approach of meta-learning or ``learning to learn''~\cite{finn2017model,8954109,rusu2018meta}.

Meta-learning process is performed on a family of tasks set on disjoint meta-training and meta-testing sets. Each task includes only a limited amount of training data which requires fast adaptability of the meta-learner. The performance of the meta-learner is evaluated on meta-testing tasks. In this paper we have modified Multi-Task Modification (MTM) approach for meta-learning from~\cite{boiarov2020simultaneous} where this algorithm was investigated on Omniglot dataset, which is a less challenging setting than those discussed in this paper. Multi-task weight optimization plays the key role in MTM. Therefore, we consider members of the family of Simultaneous Perturbation Stochastic Approximation (SPSA) methods as optimizers of multi-task weights and compare them with a gradient-based approach. To the best of our knowledge, we have adapted SPSA for Tracking method for multi-task weight optimization in the one-shot meta-learning setting for the first time. To summarize, our main contributions are as follows:
\begin{itemize}
    \item We have successfully applied our MTM to the important optimization-based method MAML~\cite{finn2017model} and the important metric-based method Prototypical Networks~\cite{NIPS2017_cb8da676}. For both of these methods, the proposed approach achieved an improvement over the original method on the main one-shot learning benchmarks.
    \item We have formulated novel SPSA for Tracking method (SPSA-Track) as a multi-task weights optimizer in MTM for meta-learning. This method demonstrated a significant performance boosting on average.
    \item We have shown robustness of SPSA-based approaches on multiple benchmarks. Also, they appear to outperform gradient-based methods as a multi-task weights optimizer.
    \item We have demonstrated that combining MTM SPSA-Track with modern backbone model achieves a significant boost in accuracy. Such a performance improvement puts our result among the best in the field.
\end{itemize}

\section{Related Works}\label{sec:related_works}


{\bf Few-Shot Learning.} Meta-learning attempts to acquire general knowledge of a target domain by learning many tasks that lie within it~\cite{finn2017model}. Few-shot learning is widely used as one of the main benchmarks for meta-learning approaches~\cite{vinyals2016matching,finn2017model,tian2020rethinking}. In one-shot learning setting, training set consists of only one sample per task. It is expected that by training on similar yet different tasks the model will learn common features that will still be relevant to unseen tasks and, thus, acquire general understanding of the field. Few-shot learning models are typically divided into 2 broad categories based on how the problem is modelled: optimization-based and metric-based.

The class of optimization-based few-shot learning algorithms uses explicit optimization for fast adaptation to new tasks. Model-Agnostic Meta-Learning (MAML)~\cite{finn2017model} attempted to find network weights that are able to quickly adapt to new tasks through an optimization procedure. Many refinements to the MAML algorithm have been proposed since its inception, for instance, Latent Embedding Optimization (LEO)~\cite{rusu2018meta} tried to improve generalization by learning in a lower-dimensional latent embedding space. R2D2~\cite{bertinetto2018metalearning} and MetaOptNet~\cite{8954109} improved accuracy by using Ridge Regression and SVM as classifiers.

Metric-based approaches are a class of methods for few-shot learning problems that aim to learn a discriminative embedding transferable to a target task. Metric learning has a long history of research and various applications~\cite{boiarov2019large,musgrave2020metric}. Siamese Convolutional Neural Networks~\cite{koch2015siamese} was the first metric-based method for one-shot image classification that learned task-agnostic feature embeddings. Matching Networks~\cite{vinyals2016matching} enhanced this approach by using two different networks and episodic training. Prototypical Networks (ProtoNet)~\cite{NIPS2017_cb8da676} introduced the idea of learning class representation by using mean features embeddings. Two other important approaches were suggested in Relational Networks~\cite{sung2018learning} that introduced an architecture to model the similarity of embeddings and Task Dependent Adaptive Metric (TADAM)~\cite{NEURIPS2018_66808e32} that switched to learning task-specific features. Recently, Deep Subspace Networks (DSN)~\cite{simon2020adaptive} have been used to predict class labels by calculating the distance between a query point and its projections onto discriminative subspaces formed by the support sets for each class.

Transductive inference as an approach to the few-shot learning problem was the subject of research in several recent papers~\cite{Dhillon2020A}. In this setting a classifier model performs class predictions jointly for all the unlabeled query examples of a single few-shot learning task, instead of making predictions for one sample at a time as in inductive methods. As an alternative to the meta-learning framework, several recent studies investigated performance of standard end-to-end pre-trained classifiers on few-shot tasks~\cite{chen2019closer,tian2020rethinking,Dhillon2020A}. However, in this work we focus on improving the meta-training phase and consider only the inductive methods that use meta-learning pipeline without pre-training. For this reason, we selected MAML and ProtoNet because papers describing these methods are among the most cited in the field. Thus, we expect that results of applying multi-task meta-learning modification should generalize when used for the modification of other one-shot learning algorithms following these restrictions.


{\bf Multi-Task Learning.} Multi-task learning is an approach where a model is trained to give predictions for multiple tasks jointly. Multi-tasking can be considered as an implicit data augmentation and regularization technique assisting the network to focus on more relevant features in the input~\cite{ruder2017overview}. However, proper task weighting might be crucial to improving model performance. Kendall {\it et al.}~\cite{kendall2018multi} have introduced a multi-task loss function that relies on maximizing the Gaussian likelihood with task-dependent uncertainty. The proposed single model has outperformed separate models for each task. In~\cite{boiarov2020simultaneous}, multi-tasking approach has been applied for a few-shot character recognition problem, which resulted in an improvement over the baseline model. A close connection in terms of optimization task formulation between multi-task learning and gradient-based meta-learning was established in~\cite{wang2021bridging}. 


{\bf Stochastic Approximation. }The stochastic approximation algorithm was developed for solving the optimization problem by Kiefer and Wolfowitz~\cite{kiefer1952stochastic}. Simultaneous Perturbation Stochastic Approximation (SPSA) algorithm~\cite{spall1992multivariate} uses only two observations at each iteration which recursively generates estimates along random directions. In high-dimensional setting SPSA has the same order of convergence rate as Kiefer-Wolfowitz approach while requiring significantly fewer measurements of a function. When an unknown but bounded disturbance corrupts the observed data, the quality of methods based on stochastic gradient decreases. However, SPSA-like algorithms demonstrate a high level of resistance to such disturbances~\cite{granichin2015randomized}. Stochastic approximation algorithms are successfully used in various machine learning problems~\cite{granichin2015randomized}.

\section{Multi-Task Meta-Learning Modification}

\subsection{One-Shot Learning Problem Definition}

We consider one-shot learning problem as a special case of few-shot learning. According to the few-shot learning problem formulation, we need to train a classifier that can quickly adapt to new unseen classes using only few labeled examples of classes. To cast this problem as meta-learning problem, Vinyals  {\it et al.}~\cite{vinyals2016matching} proposed the pipeline where elements of each class were randomly divided into {\it support set} and {\it query set}. A set of classes in training phase does not overlap with a set of classes in testing phase. In the meta-learning pipeline, training and testing processes consist of a series of episodes where each episode $\xi_t$ includes one or more tasks and each task $t_i$ consists of support and query sets for several classes.

Consider labeled dataset $\left\lbrace (\mathbf{x}_1, y_1),\ldots, (\mathbf{x}_{CN}, y_{CN})  \right\rbrace$ with $C$ classes and $N$ examples per each class, where $\mathbf{x}_i \in \mathbb{R}^d$ is the input data vector and $y_i \in \left\lbrace 1,\ldots, C \right\rbrace$ is the class label. Let $N_S$ be the number of examples in the support set for each class, $N_Q$ be the number of examples in the query set, $N_S~+~N_Q~=~N$, $N_C \leq C$ be the number of classes in a task that are randomly selected from all set of $C$ classes. Corresponding few-shot learning classification procedure is called {\it $N_S$-shot $N_C$-way classification}. In one-shot setting, $N_S=1$.

Let episode $\xi_t: (t_1, \ldots, t_M)$ consist of $M$ tasks. Each task $t_i$ 
involves per task support set $S_{t_i}=\left\lbrace (\mathbf{x}_i, y_i) \right\rbrace_{i=1}^{N_S N_C}$ and query set $Q_{t_i}=\left\lbrace (\mathbf{x}_i, y_i) \right\rbrace_{i=1}^{N_Q N_C}$ containing elements of all $N_C$ classes included in the task $t_i$ of per class support and query sets, respectively. Parameter $M \in \mathbb{N}$ can vary in different algorithms. For each task $t_i$ in an episode, $\mathcal{L}_{\theta, t_i} (D)$ denotes the value of loss function on $D$, where $\theta$ is the vector of learnable parameters of model $\phi_{\theta}$ (convolutional neural network in this paper) and $D$ can be support, query, or another dataset. 

{\bf Model-Agnostic Meta-Learning}. MAML uses loss function which is a two-step training procedure: \\1) adaptation step (inner-loop) where network parameters are adapted to the specific task $t_i$:
\begin{equation}\label{inner_loss}
	\theta'_i=\theta-\alpha\nabla_{\theta}\mathcal{L}_{\theta, t_i} (S_{t_i});
\end{equation}
2) meta-gradient update (outer-loop) step where parameters $\theta$ are updated by backpropagating through the adaptation procedure: 
\begin{equation}\label{outer_loss}
	\theta \leftarrow \theta - \beta \nabla_{\theta} \sum_{i=1}^{M} { \mathcal{L}_{\theta'_i, t_i} (Q_{t_i})},
\end{equation} 
where $\alpha$, $\beta$ are adaptation and meta-learning rates correspondingly.

{\bf Prototypical Networks}. In ProtoNet algorithm model $\phi_{\theta}$ computes embedding of input data. Each class $k$ is represented by prototype vector $\mathbf{c}^k_{t_i}$ which is calculated as a mean vector of the corresponding support set. The Prototypical Networks model $\phi_{\theta}$ is trained via stochastic gradient descent (SGD) by minimizing loss function for train task $t_i$, where $d(\cdot, \cdot)$ is Euclidean distance function:
\begin{equation}
\mathcal{L}_{\theta, t_i} (Q_{t_i}) = \frac{1}{N_C} \sum_{k=1}^{N_C} \frac{1}{N_Q} \sum_{\mathbf{x}_j \in Q_{t_i}^k} -\log \frac{\exp(-d(\phi_\theta (\mathbf{x}), \mathbf{c}^k_{t_i}))}{\sum_{k'} \exp(-d(\phi_\theta (\mathbf{x}), \mathbf{c}^{k'}_{t_i}))}.
\end{equation}
\begin{figure*}[t]
  \centering
  \includegraphics[width=0.8\textwidth]{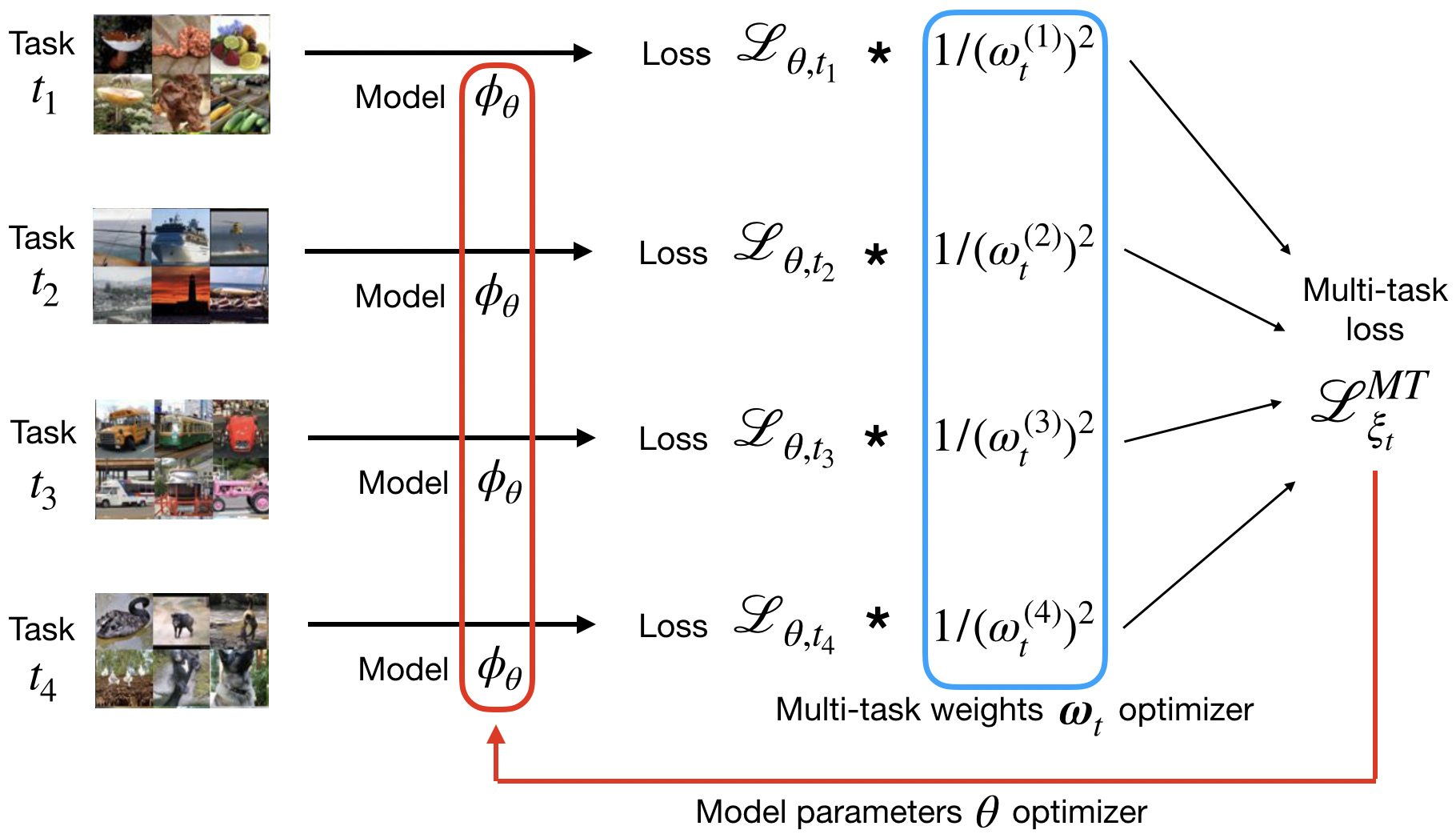}
  \caption{Multi-task meta-learning modification pipeline for a training episode $\xi_t$ described in Algorithm~1.}
  \label{fig:mtm_pipeline}
\end{figure*}

\subsection{Multi-Task Meta-Learning Loss Function}

Many few-shot learning methods~\cite{vinyals2016matching,NIPS2017_cb8da676,8954109} work with $M=1$ (i.e. with one task in a training episode $\xi_t$). Other approaches~\cite{finn2017model,mishra2017simple} utilize a batch of tasks with $M > 1$ per episode but consider the contribution of each task equally. In this paper we develop the idea of adaptive use of information from multiple tasks simultaneously via the multi-task approach for meta-learning proposed in~\cite{boiarov2020simultaneous}.

Multi-task learning methods for deep neural networks can be divided into two main areas: soft and hard parameter sharing of hidden layers of neural network~\cite{ruder2017overview}. In accordance with~\cite{boiarov2020simultaneous}, we use hard parameter sharing for all hidden layers of a convolutional network. Thus, one neural network is used for all tasks, and the presence of several tasks is reflected only in the loss function. For this purpose~\cite{boiarov2020simultaneous} proposed an adapted approach discussed in~\cite{kendall2018multi} which used task-depended (homoscedastic) uncertainty as a basis for weighting losses in a multi-task learning problem. Corresponding multi-task meta-learning loss function takes the following form:

\begin{equation}\label{multitask_loss}
\mathcal{L}^{MT}_{\xi_t}(\boldsymbol{\omega}_t, \left\lbrace Q_{t_i} \right\rbrace_{i=1}^{M}) = \sum_{i=1}^M \frac{1}{(\omega_t^{(i)})^2} \mathcal{L}_{\theta, t_i} (Q_{t_i}) + \sum_{i=1}^M \log (\omega_t^{(i)})^2,
\end{equation}
where weights $\boldsymbol{\omega}_t = (\omega_t^{(1)},\ldots,\omega_t^{(M)})$ are hyperparameters.

\subsection{Multi-Task Weights Optimization}\label{sec:mtm_optimizers}

As shown in~\cite{kendall2018multi}, model performance is extremely sensitive
to multi-task weights selection and tuning of $\boldsymbol{\omega}_t$ is critical for the success of a multi-task learning. On the other hand, searching for these optimal weights is expensive and increasingly difficult for a large model with numerous tasks.

In this work we focus on developing and investigating performance of optimization methods for hyperparameters $\boldsymbol{\omega}_t$ in the loss function~(\ref{multitask_loss}). Nowadays gradient optimization approach is a natural choice in many deep learning algorithms~\cite{He_2016_CVPR,kendall2018multi}. Therefore, we consider an approach with embedding weights optimization in the backpropagation procedure similar to the one described in~\cite{kendall2018multi}. This procedure is described in Algorithm~1. Let $\widehat{\boldsymbol{\omega}}_{t}$ and $\widehat{\theta}_{t}$ denote estimates at iteration $t$ of multi-task weights and model parameters, respectively. This multi-task meta-learning modification pipeline for a training episode $\xi_t$ is visualized in Figure~\ref{fig:mtm_pipeline}.

\begin{algorithm}\label{alg:multi-task_weights_optimizer}
\begin{algorithmic}[1]
\caption{Training for episode $\xi_t: (t_1,\ldots,t_M)$}
\renewcommand{\algorithmicrequire}{\textbf{Input:}}
\renewcommand{\algorithmicensure}{\textbf{Output:}}
\REQUIRE $N_S$, $N_Q$, $N_C$, $\widehat{\theta}_{t-1}$, $\widehat{\boldsymbol{\omega}}_{t-1}$
\ENSURE Updated parameters estimates $\widehat{\theta}_{t}$, $\widehat{\boldsymbol{\omega}}_{t}$
\FOR{$i$ in $\left\lbrace 1,\ldots,M \right\rbrace$}
\STATE Sample $N_C$ random classes
\STATE Sample random elements in $S_{t_i}$ and $Q_{t_i}$ 
\STATE Compute task loss function $\mathcal{L}_{\theta, t_i} (Q_{t_i})$
\ENDFOR
\STATE Compute $\mathcal{L}^{MT}_{\xi_t}(\widehat{\boldsymbol{\omega}}_{t-1}, \left\lbrace Q_{t_i} \right\rbrace_{i=1}^{M})$ via~(\ref{multitask_loss})
\STATE Use {\bf multi-task weights optimizer} to update $\widehat{\boldsymbol{\omega}}_{t}$
\STATE Update parameters $\widehat{\theta}_{t}$ via SGD by $\mathcal{L}^{MT}_{\xi_t}$
\end{algorithmic}
\end{algorithm}


In this paper gradient-based and stochastic approximation methods are considered and compared as the multi-task weights optimizer. 

When solving a meta-learning problem, algorithms face model uncertainties associated with a critical issue of limited amount of training samples per task, which may result in overfitting~\cite{nguyen2020uncertainty}. This phenomenon is especially pronounced in the extreme case of one-shot learning. Therefore, meta-learning algorithms must be able to effectively adapt to these uncertainties~\cite{finn2018probabilistic, abdar2021review}. We approach this issue from optimization point and propose to use methods from the family of Simultaneous Perturbation Stochastic Approximation (SPSA) algorithms as a multi-task weights optimizer due to their robustness and successful application in various machine learning and control problems with uncertainties~\cite{yue2007using, granichin2015randomized, boiarov2019stochastic}.


{\bf Gradient-Based Methods.} As stated earlier, one of the proposed gradient-based approaches is to incorporate multi-task weights in a computational graph of meta-learning algorithm (MAML and ProtoNet in this work). In that case backpropagation procedure is used to compute weights updates and a gradient-based optimizer is shared with the whole model. This multi-task modification of meta-learning is denoted by {\it MTM Backprop}. We also explored first-order optimization approaches as a separate multi-task weights optimizer described in Algorithm~1. Adam optimization method~\cite{kingma2015adam} was selected based on the results of experiments and due to multiple mentions in the deep learning literature. This approach is named {\it MTM Inner First-Order} by analogy with inner loop optimization step in MAML~(\ref{inner_loss}).



\subsection{SPSA for Tracking method}

When solving a meta-learning problem, algorithms face model uncertainties associated with a critical issue of limited amount of training samples per task, which may result in overfitting~\cite{nguyen2020uncertainty}. This phenomenon is especially pronounced in the extreme case of one-shot learning. Therefore, meta-learning algorithms must be able to effectively adapt to these uncertainties~\cite{finn2018probabilistic,abdar2021review}.
We approach this issue from optimization point and propose to use methods from the family of SPSA algorithms as a multi-task weights optimizer due to their robustness and successful application in various machine learning and control problems with uncertainties~\cite{granichin2015randomized}. In this subsection, we propose the SPSA for Tracking algorithm firstly adapted for multi-task weights optimization.

In order to cast the SPSA-based approach as a multi-task weights optimizer, we reformulate the problem of finding estimates of hyperparameters $\boldsymbol{\omega}_t$ in the multi-task loss function~(\ref{multitask_loss}) as a non-stationary optimization problem suitable for a stochastic approximation~\cite{granichin2014simultaneous,boiarov2020simultaneous}. In order to do this, we introduce an observation model for the training episode~$\xi_t$ that takes into account uncertainties arising in meta-learning:
\begin{equation}\label{non_stationar_opt_func}
L_t(\boldsymbol{\omega}_t) = \mathcal{L}^{MT}_{\xi_t}(\boldsymbol{\omega}_t, \left\lbrace Q_{t_i} \right\rbrace_{i=1}^{M}) + \nu_t,
\end{equation}
where $\nu_t$ is an additive external noise caused by uncertainties arising from the limited amount of training samples per task. Therefore, it is necessary to find an algorithm that produces an estimate $\widehat{\boldsymbol{\omega}}_t$ of an unknown vector $\boldsymbol{\omega}_t$ that minimizes mean-risk functional of objective functions~(\ref{non_stationar_opt_func}) based on observations $L_1, L_2,\ldots,L_t$ from training episodes $\xi_1, \xi_2,\ldots,\xi_t$.

To the best of our knowledge, we propose to use SPSA for Tracking approach~\cite{granichin2014simultaneous} as a multi-task weights optimizer for the first time. This algorithm simultaneously uses observation from the current and previous iterations which allows it to be more stable under parameters drift conditions. Let $\boldsymbol{\Delta}_t \in {\mathbb R}^d$ be a vector consisting of independent random variables with Bernoulli distribution, $\widehat{\boldsymbol{\omega}}_{0}$ a vector with the initial values, and $\{\alpha_t\}$ and $\{\beta_t\}$ sequences of positive numbers. Then SPSA for tracking multi-task weights optimizer ({\it MTM~SPSA-Track}) constructs the following estimates of multi-task weights: 

\begin{eqnarray}\label{eq:fsl_spsa_track}
\begin{cases}
L_{2t} = L_{2t}(\widehat{\boldsymbol{\omega}}_{2t-2} + \beta_n \boldsymbol{\Delta}_t), \;\\ L_{2t-1} = L_{2t-1}(\widehat{\boldsymbol{\omega}}_{2t-2} - \beta_t \boldsymbol{\Delta}_t)
\\
\\
\widehat{\boldsymbol{\omega}}_{2t-1} = \widehat{\boldsymbol{\omega}}_{2t-2}
\\
\widehat{\boldsymbol{\omega}}_{2t} = \widehat{\boldsymbol{\omega}}_{2t-1} - \alpha_t \boldsymbol{\Delta}_t \frac{L_{2t} - L_{2t-1}}{2\beta_t}.
\end{cases}
\end{eqnarray}

Algorithm~(\ref{eq:fsl_spsa_track}) has theoretical convergence guarantees expressed in terms of an upper bound of residuals between the estimates and the theoretical optimal solution~\cite{granichin2014simultaneous}.
\subsection{Multi-Task Modification for Class-Based Weights}

In Algorithm~1, $N_C$ classes in task $t_i$ is sampled randomly. However, if additional meta-information about classes is available, the data can be sampled from some coarse classes. For this, we define the modified multi-task loss function~(\ref{multitask_loss}) applicable to coarse classes as well as to usual finer-grained classes. A per-task loss function $\mathcal{L}_{\theta, t_i}$ can be rewritten as a sum of terms corresponding to classes:
\begin{equation*}
\mathcal{L}_{\theta, t_i} (Q_{t_i}) = \sum_{k=1}^{C} \mathcal{L}_{\theta, t_i, k} (Q_{t_i}),
\end{equation*}
where $k$ is a class, $k \in \left\lbrace 1,\ldots, K \right\rbrace$, $K$ is the number of coarse classes (or $K = C$), $\mathcal{L}_{\theta, t_i, k} (Q_{t_i})$ is the algorithm-specific partial loss computed for the corresponding class which is equal to zero if the class $k$ is not present in the task $t_i$. Then we modify multi-task loss function~(\ref{multitask_loss}) by introducing weights $\omega_C^{(k)}$ for each class and combining like terms from $M$ training tasks:
\begin{equation}\label{eq:coarse_multitask_loss}
\mathcal{L}^{MT}_{\xi_t}(\boldsymbol{\omega}_t, \left\lbrace Q_{t_i} \right\rbrace_{i=1}^{M}) = \sum_{k=1}^C \left( \left( \frac{1}{(\omega_C^{(k)})^2} \sum_{i=1}^{M} \mathcal{L}_{\theta, t_i, k} (Q_{t_i}) \right) +  \log (\omega_C^{(k)})^2\right).
\end{equation}
SPSA-based multi-task weights optimizer with loss function~(\ref{eq:coarse_multitask_loss}) is denoted as {\it MTM SPSA-Coarse}.

\section{Experiments}

We conducted experiments on four datasets -- CIFAR-FS~\cite{bertinetto2018metalearning}, FC100~\cite{NEURIPS2018_66808e32}, miniImageNet~\cite{vinyals2016matching} and tieredImageNet~\cite{ren2018metalearning}. These datasets were selected as they became standard benchmarks over the last couple of years~\cite{Dhillon2020A, tian2020rethinking}. Experiments were performed on Rolos platform\footnote[1]{https://rolos.com/} with a A100 vGPU, 48 vCPUs at 2.30GHz, 128GiB RAM. Tables~\ref{table:all_benchmarks},~\ref{table:grad_vs_spsa} and~\ref{table:all-mini-imagenet} summarize results of these experiments.

The CIFAR-FS and FC100 datasets are derived from CIFAR-100 dataset having 100 classes with each class consisting of 600 images of size 32~×~32. In the CIFAR-FS dataset classes are randomly divided into groups of 64, 16 and 20 for training, validation, and testing, respectively, while in the FC100 datasets 20 superclasses of CIFAR-100 are split into groups of 12, 4 and 4. The miniImageNet and tieredImageNet datasets are derived from ILSVRC-2012 dataset having 1000 categories. The miniImageNet dataset consists of 100 classes randomly chosen from ILSVRC-2012 split into groups of 64, 16 and 20 as proposed in~\cite{Ravi2017OptimizationAA} which is a standard convention. The tieredImageNet dataset consists of 608 categories from ILSVRC-2012 grouped into 34 supercategories that are then split into  groups of 20, 6 and 8. In both cases 600 images of 84~×~84 pixels in size are sampled for each class.

\begin{table*}[th]
	\caption{Multi-Task Modification results on CIFAR-FS, FC100, miniImageNet and tieredImageNet (1-shot setting). Improvements are shown in {\bf bold}.}
	\label{table:all_benchmarks}
	\begin{center}
		\begin{tabular}{|l|c|c|c|c|c|c|c|c|}
	    	\hline
		     & \multicolumn{2}{|c|}{CIFAR-FS} & \multicolumn{2}{c|}{FC100} & \multicolumn{2}{|c|}{miniImageNet} & \multicolumn{2}{c|}{tieredImageNet}\\
			\hline
			Configuration & 2-way & 5-way & 2-way & 5-way & 2-way & 5-way & 2-way & 5-way \\
			\hline
			\multicolumn{9}{|c|}{MAML} \\
			\hline
			Reproduced     & \,74.8\,\%    & \,54.5\,\%    & \,66.0\,\%    & \,36.4\,\%    & \,73.2\,\%    & \,45.9\,\%    & \,73.3\,\%    & \,47.9\,\%   \\
			MTM SPSA       & \bf{76.4\,\%} & \bf{54.7\,\%} & \bf{67.0}\,\% & 36.4\,\%      & \bf{74.9\,\%} & \bf{48.0\,\%} & \bf{74.9}\,\% & 47.8\,\% \\
			MTM SPSA-Track & \bf{76.8\,\%} & \bf{54.6\,\%} & \bf{66.8}\,\% & \bf{38.3\,\%} & \bf{75.8\,\%} & \bf{46.5\,\%} & \bf{73.8}\,\% & \bf{48.3\,\%} \\
			\hline
			\multicolumn{9}{|c|}{ProtoNet} \\
			\hline
			Reproduced     & \,77.8\,\%    & \,58.9\,\%    & \,65.0\,\%    & \,35.7\,\%    & \,74.2\,\%    & \,50.0\,\%    & \,72.9\,\%    & \,49.4\,\%   \\
			MTM SPSA       & \bf{79.1\,\%} & \bf{59.7\,\%} & \bf{65.1}\,\% & \bf{36.0\,\%} & \bf{74.7\,\%} & \bf{50.2\,\%} & \bf{73.6}\,\% & \bf{49.5\,\%} \\
			MTM SPSA-Track & \bf{78.2\,\%} & \bf{59.8\,\%} & \bf{65.3}\,\% & \bf{36.1\,\%} & \bf{74.8\,\%} & \bf{50.8\,\%} & \bf{73.1}\,\% & \bf{50.0\,\%} \\
			\hline
		\end{tabular}
	\end{center}
\end{table*}

As mentioned in Section~\ref{sec:related_works}, the experiments with Multi-Task Modification were performed using MAML and Prototypical Networks. For MAML algorithm we used the neural network $\phi_{\theta}$ defined in the original paper~\cite{finn2017model} with 32-32-32-32 configuration where a-b-c-d denotes a 4-layer convolutional neural network with a, b, c, d filters in convolutional layers. Adam was used as the meta-optimizer with learning rate $\beta = 10^{-3}$ and adaptation step size $\alpha = 0.01$ as in~(\ref{outer_loss}). We used 5 adaptation steps during training and 10 during testing for all datasets. We selected meta-batch size of $M = 4$ and trained the model for 300 epochs, each containing 100 tasks, unless specified otherwise. Following~\cite{Ravi2017OptimizationAA} and the original paper, 15 samples per class were taken for evaluation. For the MTM the baseline model was trained for extra 40 epochs.

In the case of ProtoNet we used 64-64-64-64 feature extraction backbone $\phi_{\theta}$ as suggested in the original paper~\cite{NIPS2017_cb8da676}. We followed the standard practice~\cite{gidaris2018dynamic} for the meta-learning setup using SGD with Nesterov momentum of 0.9 and weight decay of 0.0005 as an optimizer. The learning rate was initially set to 0.1 and then decreased according to the strategy from~\cite{gidaris2018dynamic}. During meta-training, we used the first 20 epochs for pre-training the model by using the original ProtoNet method and implementing the Multi-Task Modifications only for the last 40 epochs. 

We designed several experiment settings to research the relative advantage of using the multi-task loss function~(\ref{multitask_loss}) and SPSA-based optimization over original methods (Table~\ref{table:all_benchmarks}) and over gradient-based method (Table~\ref{table:grad_vs_spsa}) where multi-task weights in the loss function are optimized jointly with the network parameters~$\theta$. In experiments we mainly used $M = 4$ tasks per training episode $\xi_t$ (other values of $M$ are indicated explicitly) and the best model for testing was selected on the validation set. For SPSA and SPSA-Track we set $\alpha_n=0.25 / n^{\frac{1}{6}}, \beta_n~=~15 / n^{\frac{1}{24}}$ in~(\ref{eq:fsl_spsa_track}) as per the theoretical result from~\cite{granichin2015randomized}. During the experiments we found that $L^2$-normalization of multi-task weights in MTM SPSA and MTM SPSA-Track improves the stability of training. Results of all experiments are formulated in terms of average few-shot classification accuracy after 1000 testing iterations.

\begin{table}[th]
	\caption{Comparison of gradient-based (Backprop), SPSA and SPSA for Tracking multi-task weights optimizers on 1-shot, 5-way experiments.}
	\label{table:grad_vs_spsa}
	\begin{center}
		\begin{tabular}{|l|c|c|c|c|}
	    	\hline
		    Algorithm & \; CIFAR-FS \; & \; FC100 \; & \; miniImageNet \; & \; tieredImageNet \; \\
		    \hline
		    \multicolumn{5}{|c|}{MAML} \\
			\hline
			Backprop   & 53.1 \% & 37.6 \% & 47.4 \% & 46.7 \%  \\
			SPSA         & \bf{54.7 \%} &  36.4 \% & \bf{48.0 \%} & 47.8 \%   \\
			SPSA-Track \;   & 54.6 \% & \bf{38.3 \%} & 46.5 \% & \bf{48.3 \%}   \\
			\hline
			 \multicolumn{5}{|c|}{ProtoNet} \\
			\hline
			Backprop   & 59.4 \% & 35.5 \% & 50.4 \% & 49.2 \%  \\
			SPSA   & 59.7 \% & 36.0 \% & 50.2 \% & 49.5 \%  \\
			SPSA-Track   & \bf{59.8 \%} & \bf{36.1 \%} & \bf{50.8 \%} & \bf{50.0 \%}  \\
			\hline
		\end{tabular}
	\end{center}
\end{table}

We used the miniImageNet dataset to compare our method with the prior work. Since the majority of recent approaches use more advanced convolutional neural networks with higher embedding dimension such as residual networks (ResNet)~\cite{He_2016_CVPR} as feature extraction backbones, we implemented the original ProtoNet with ResNet-12 backbone provided in~\cite{8954109} to compare against the results of other methods with backbones from the ResNet family. We did not include approaches that were developed for semi-supervised and transductive learning settings in this comparison since such approaches use the statistics of query examples or statistics across the one-shot tasks. We also excluded methods that use non-episodic pre-training as mentioned in Section~\ref{sec:related_works}. Table~\ref{table:all-mini-imagenet} shows that, when used with a comparable backbone, our multi-task meta-learning modification of ProtoNet with stochastic approximation increases one-shot classification accuracy to the level of significantly more advanced methods and is competitive against state-of-the-art meta-learning approaches. These results are presented with 95~\% confidence intervals.

\begin{table*}[th]
	\caption{Comparison to prior work on miniImageNet meta-test split. {\bf Bold} values are the accuracy no less than 1 \% compared with the highest one.} 
	\label{table:all-mini-imagenet}
	\begin{center}
		\begin{tabular}{|p{6.5 cm}|c|c|}
			\hline
			Algorithm & Backbone & 1-shot 5-way \\
			\hline
 			MAML~\cite{finn2017model,chen2019closer} & \; ResNet-18 \; & 49.61 $\pm$ 0.92 \% \\
 			Chen et al.~\cite{chen2019closer} & ResNet-18 & 51.87 $\pm$ 0.77 \%  \\
 			Relation Networks~\cite{sung2018learning,chen2019closer} &  ResNet-18 & 52.48 $\pm$ 0.86 \%  \\
 			Matching Networks~\cite{vinyals2016matching,chen2019closer} & ResNet-18 & 52.91 $\pm$ 0.88 \%  \\
			RAP-ProtoNet~\cite{hong2021reinforced} & ResNet-10 & 53.64 $\pm$ 0.60 \% \\
 			ProtoNet (reproduced)~\cite{NIPS2017_cb8da676} & ResNet-12 & 56.52 $\pm$ 0.45 \%  \\
 			Gidaris  {\it et al.}~\cite{gidaris2018dynamic} & ResNet-15 & 55.45 $\pm$ 0.89 \% \\
			SNAIL~\cite{mishra2017simple} & ResNet-15 &  55.71 $\pm$ 0.99 \% \\
			Bauer  {\it et al.}~\cite{bauer2017discriminative} & ResNet-34 &  56.30 $\pm$ 0.40 \%  \\
			AdaResNet~\cite{munkhdalai2018rapid} & ResNet-12 &  56.88 $\pm$ 0.62 \%  \\
			TADAM~\cite{NEURIPS2018_66808e32} & ResNet-12 &  58.50 $\pm$ 0.30 \% \\
			Shot-Free~\cite{ravichandran2019few} & ResNet-12 &  59.04 $\pm$ n/a \%  \\
			CAML~\cite{jiang2018learning} & ResNet-12 &   59.23 $\pm$ 0.99 \%  \\
			Wang {\it et al.}~\cite{wang2021bridging} & ResNet-12 &   59.84 $\pm$ 0.22 \%  \\
			MTL~\cite{sun2019meta} & ResNet-12 & 61.20 $\pm$ 1.80 \% \\
			vFSL~\cite{zhang2019variational} & ResNet-12 & 61.23 $\pm$ 0.26 \%  \\
			MetaOptNet~\cite{8954109} & ResNet-12 &  \bf{62.64 $\pm$ 0.61 \%}  \\
			DSN~\cite{simon2020adaptive} & ResNet-12 &  \bf{62.64 $\pm$ 0.66} \%  \\
			\hline
			ProtoNet MTM SPSA (Ours) & ResNet-12 & 60.24 $\pm$ 0.71 \%  \\
			ProtoNet MTM SPSA-Track (Ours) & ResNet-12 & 61.33 $\pm$ 0.74 \%   \\
			ProtoNet MTM SPSA-Track ($M=2$) (Ours) & ResNet-12 & \; \bf{61.94 $\pm$ 0.73 \%} \;  \\
			\hline
		\end{tabular}
	\end{center}
\end{table*}

\section{Ablation Study}

\subsection{Improvements of MTM with SPSA and SPSA-Track.} 

We have conducted experiments for four datasets most widely used in the field of one-shot learning as shown in Table~\ref{table:all_benchmarks}. On CIFAR-FS, we have improved on original methods up to ${\bf 2.0}${\bf \%}, with the largest improvement by 2-way MAML MTM SPSA-Track. On FC100 benchmark, the largest improvement of $\bf{1.9}${\bf \%} has been achieved with the novel MTM SPSA-Track method on MAML in 5-way scenario. On tieredImageNet, we got improvements up to $\bf{1.6}${\bf \%} for MAML in 2-way scenario with SPSA. The top performing methods include MTM SPSA and MTM SPSA-Track for both MAML and ProtoNet. MTM SPSA-Track gives the greatest boost in 5-way settings. The last dataset we considered was miniImageNet, which is the most widely used benchmark for few-shot learning. Here we have achieved significant improvements up to $\bf{2.6}${\bf \%} with MAML MTM SPSA-Track leading and MAML MTM SPSA following. Similar results are observed for ProtoNet. The results presented in Table~\ref{table:all_benchmarks} show that proposed MTM with SPSA and SPSA-Track methods outperform the original approaches on all four benchmarks. Novel SPSA-Track demonstrates the largest improvement over the baseline in~most~cases.

The experiment results shown in Table~\ref{table:all-mini-imagenet} suggest that applying our method to ProtoNet with a more modern ResNet-12 backbone gives performance improvement of $\bf{3.72}${\bf \%} for MTM SPSA, $\bf{4.81}${\bf \%} for MTM SPSA-Track, making MTM SPSA-Track better performing in this scenario as well. Such a performance improvement puts our result among the best in the field. Applying MTM SPSA-Track to ProtoNet with number of tasks $M=2$ gives a performance improvement of $\bf{5.42}${\bf \%} that is competitive against state-of-the-art methods. The fact that this improvement has been achieved by modifying the loss function only and the fact that MTM SPSA and MTM SPSA-Track can be applied to almost any of the meta-learning methods in the table makes our result even more significant. It is worth noting that in the case of meta-learning methods like MAML that originally use meta-batch with several tasks, our Multi-Task Modification requires practically no additional computational costs.

\subsection{Gradient-Based vs SPSA-Based Optimization.} 

We explored the comparison between gradient-based and SPSA-based approaches of multi-task weight optimization. As can be seen from Table~\ref{table:grad_vs_spsa}, SPSA-based approaches give superior results in all experiments, so we can conclude that zero-order optimizers are better suited for the proposed multi-task modification in one-shot setting.

\subsection{SPSA Multi-Task Weights Behavior}

\begin{figure}[ht]
    \centering
    \subfigure[]{\includegraphics[width=0.45\textwidth]{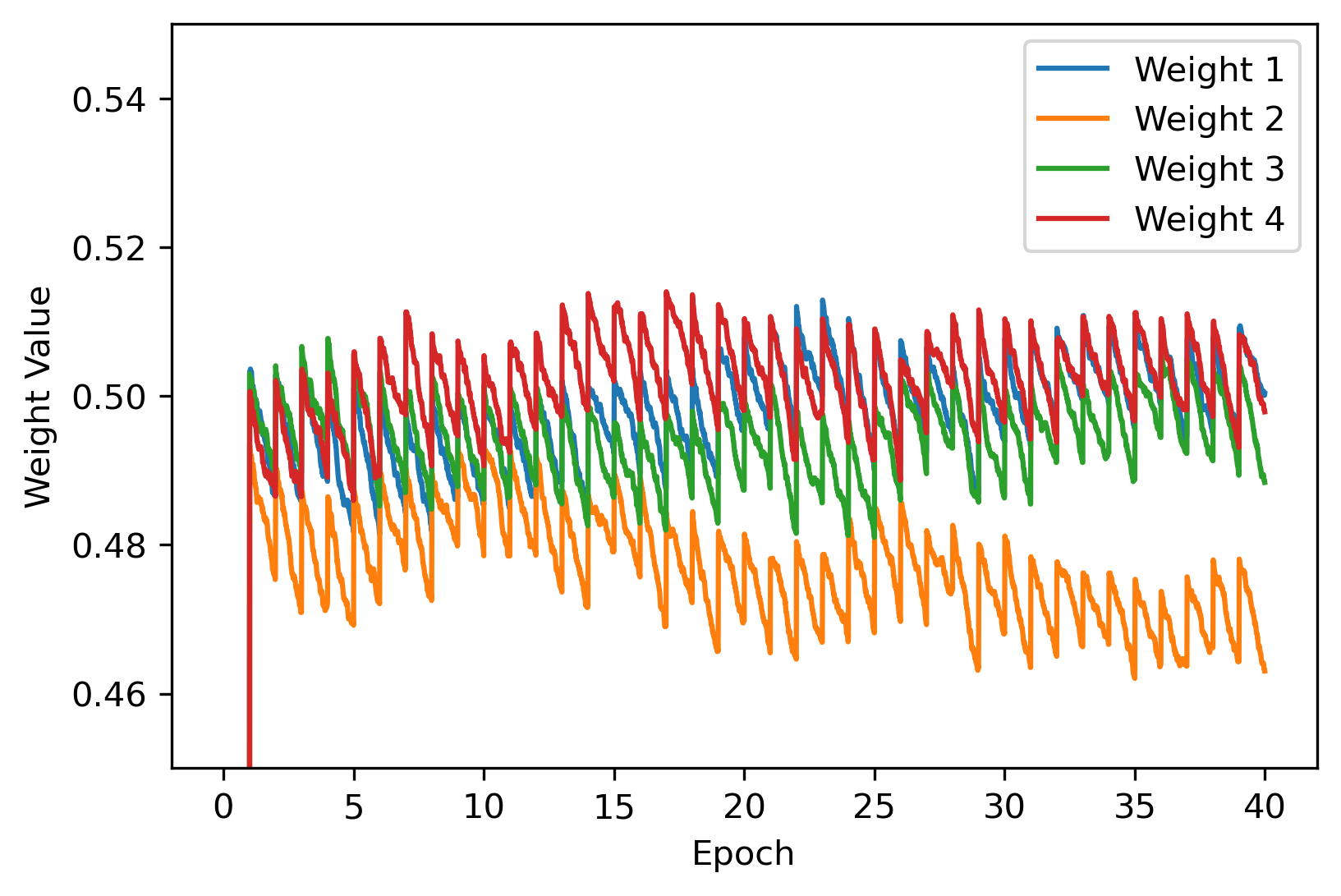}} 
    \subfigure[]{\includegraphics[width=0.45\textwidth]{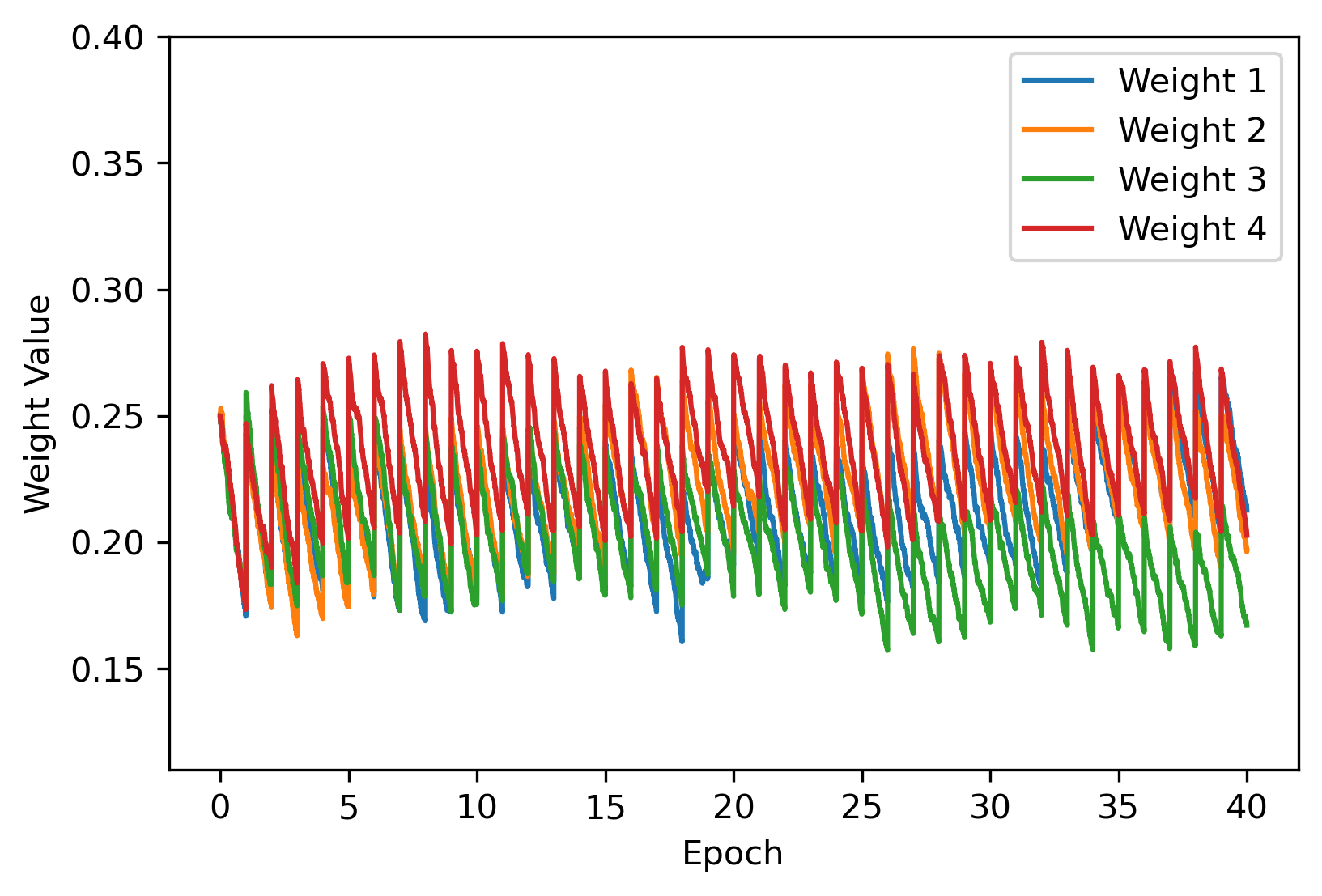}} 
    \caption{MTM SPSA-Track multi-task weights dynamic during training with weights normalization on\\ miniImageNet 1-shot 2-way: (a) MAML, (b) ProtoNet.}
    \label{fig:weight-normalization-weights}
\end{figure}

In MTM SPSA and MTM SPSA-Track experiments multi-task weights are renormalized during the optimization procedure after each epoch. The corresponding dynamics of SPSA-Track multi-task weights during training in miniImageNet 1-shot 2-way experiment is depicted in Figure~\ref{fig:weight-normalization-weights}~(a) for MAML and (b) for ProtoNet. Weights in ProtoNet experiment demonstrate larger fluctuation, yet in both cases there is a visible trend of multi-task weights splitting despite the difference in meta-learning methods.

\subsection{Task Data Sampling}

We used data sampling provided by Torchmeta library~\citep{DBLP:journals/corr/abs-1909-06576} to supply a unified approach to the problem. It follows the usual procedure where classes are sampled from the set of candidates and then an appropriate number of examples are selected per class. This procedure is randomized for every task and is independent from the choice of algorithm or its multi-task modification.

In MTM SPSA-Coarse experiment setting the tasks are sampled in the same way, however, information about coarse classes is passed to the MTM SPSA algorithm. We have modified MTM SPSA, so that instead of having $M = 4$ multi-task weights (one per task), we use 20 (one per CIFAR-100 coarse class). Consequently, instead of weighting tasks, we now weight instances of the corresponding coarse classes. A tie between a particular coarse class and the corresponding weight is established via modified multi-task loss function~(\ref{eq:coarse_multitask_loss}). At the end of a training episode the weights are updated only for the coarse classes that are present in the episode's meta-batch.

\begin{figure}[ht]
    \centering
    \subfigure[]{\includegraphics[width=0.45\textwidth]{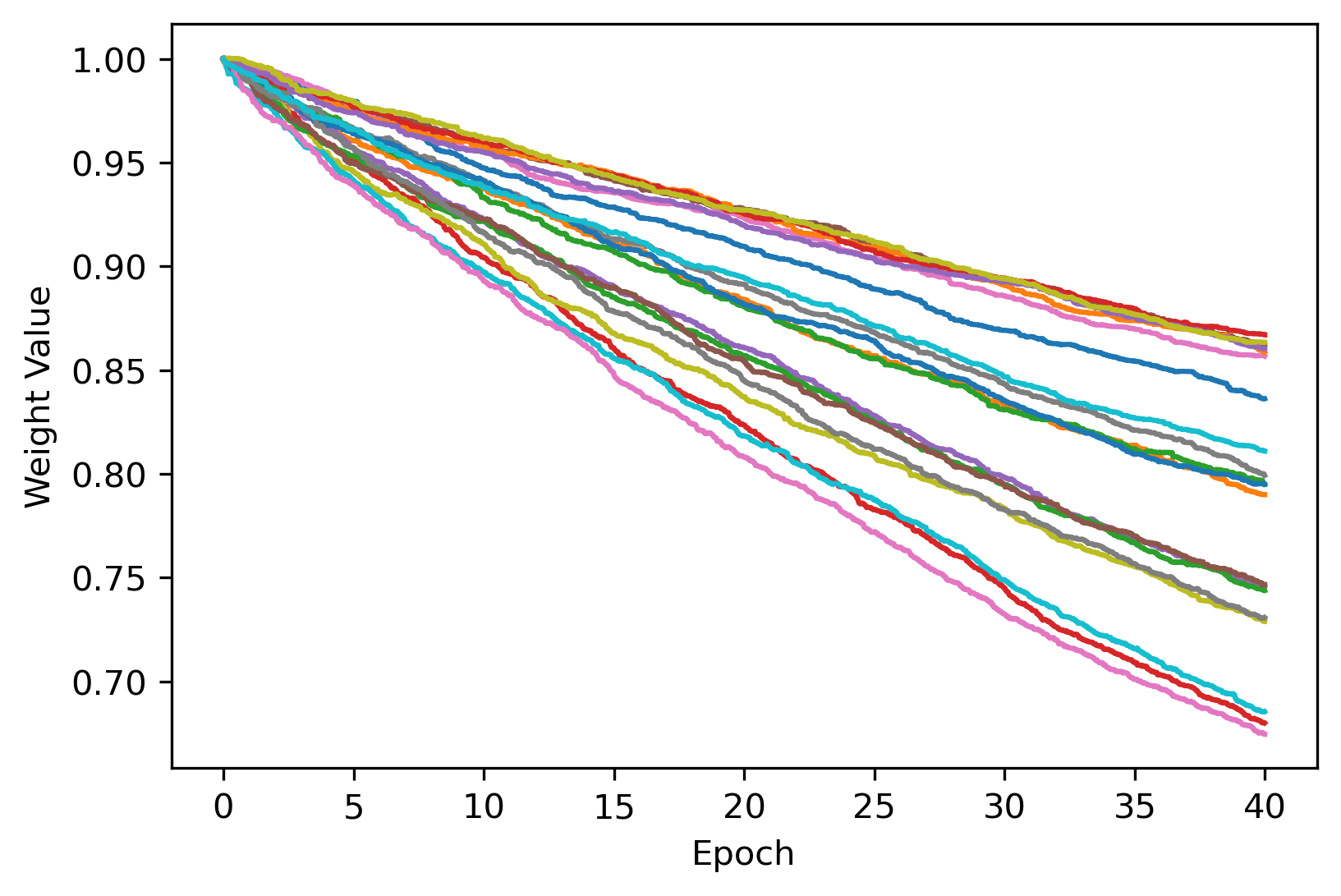}} 
    \subfigure[]{\includegraphics[width=0.45\textwidth]{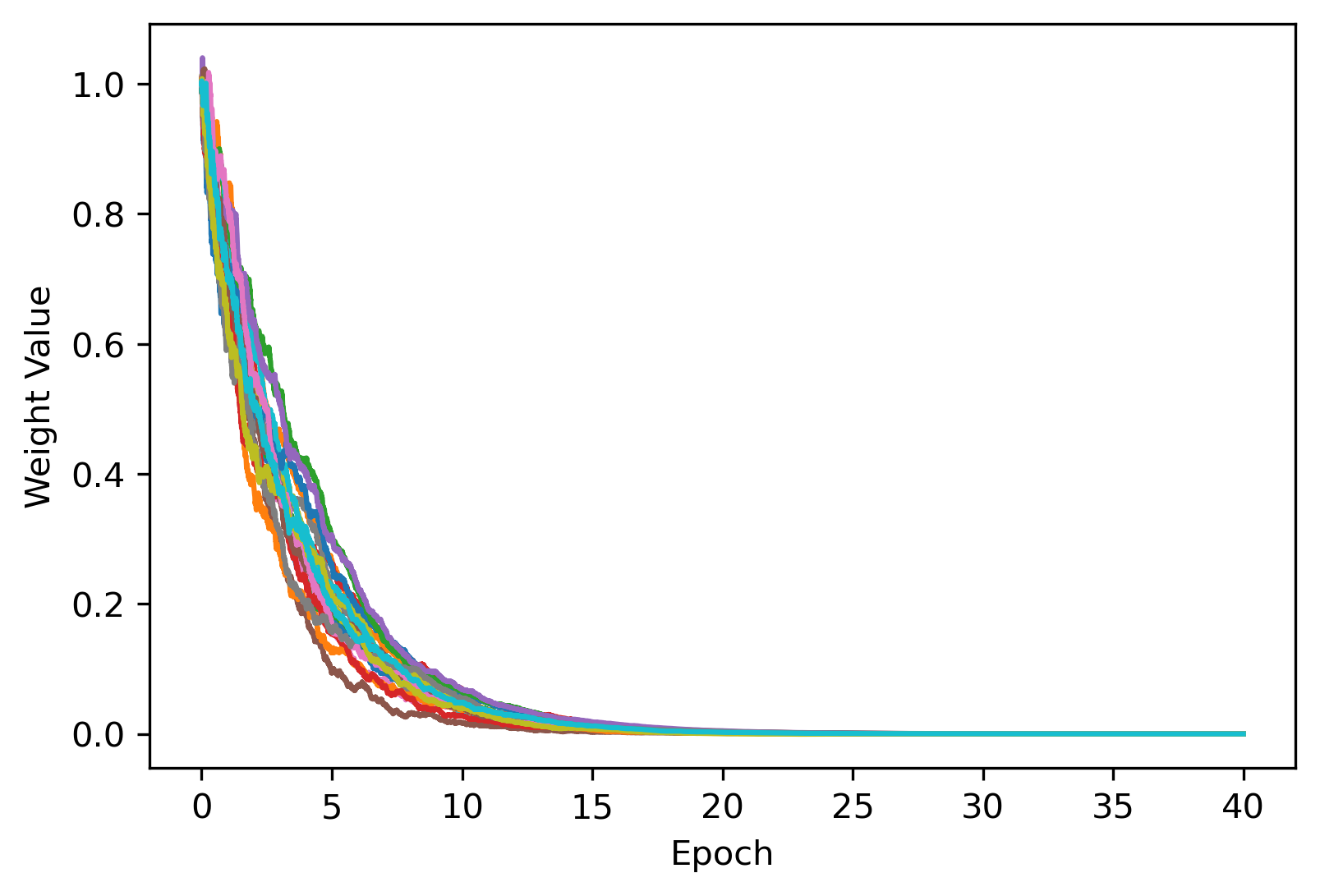}} 
    \caption{MTM SPSA-Coarse class weights dynamic during training on CIFAR-FS 1-shot 5-way:\\ (a) MAML, (b) ProtoNet.}
    \label{fig:coarse-classes-cifarfs-5way-5shot}
\end{figure}

As can be seen from Figure~\ref{fig:coarse-classes-cifarfs-5way-5shot}~(a), in case of MAML the coarse classes weights $\omega_C^{(k)}$ dynamic during training varies significantly between different classes, which contrasts with ProtoNet, where weights follow the general dynamics established in~\cite{boiarov2020simultaneous} as shown in Figure~\ref{fig:coarse-classes-cifarfs-5way-5shot}~(b). 

\section{Conclusion}

In this paper we have developed a new multi-task meta-learning modification with stochastic approximation for one-shot learning. The application of this approach to optimization-based method MAML and metric-based method ProtoNet was investigated on four most widely used one-shot learning benchmarks. In all experiments our algorithm showed significant improvements over baseline methods. In addition, in most cases SPSA-based approach was better than gradient method. We presented novel SPSA for Tracking algorithm as a multi-task weights optimizer which has demonstrated the largest performance boost on average. For future work, we aim to apply the described approach to state-of-the-art few-shot learning algorithms and to reinforcement learning.


\bibliographystyle{unsrtnat}
\bibliography{biblio}  






\end{document}